\def\year{2020}\relax
\documentclass[letterpaper]{article} 
\usepackage{aaai20}  
\usepackage{times}  
\usepackage{helvet} 
\usepackage{courier}  
\usepackage[hyphens]{url}  
\usepackage{graphicx} 
\usepackage{graphicx}  

\usepackage{subcaption}
\usepackage{amsmath}
\usepackage{booktabs} 
\usepackage{algorithm}
\usepackage{algorithmic}
\usepackage{comment}
\usepackage{appendix}

\title{How Should an Agent Practice?}
\author{Janarthanan Rajendran, Richard Lewis, Vivek Veeriah, Honglak Lee, Satinder Singh\\
University of Michigan\\
rjana@umich.edu, rickl@umich.edu, vveeriah@umich.edu, honglak@eecs.umich.edu, baveja@umich.edu}
 
\begin{document}

\maketitle

\begin{abstract}
We present a method for learning intrinsic reward functions to drive the learning of an agent during periods of {\em practice} in which extrinsic task rewards are not available. During practice, the environment may differ from the one available for training and evaluation with extrinsic rewards. We refer to this setup of alternating periods of practice and objective evaluation as {\em practice-match}, drawing an analogy to regimes of skill acquisition common for humans in sports and games. The agent must effectively use periods in the practice environment so that performance improves during matches. In the proposed method the intrinsic practice reward is learned through a meta-gradient approach that adapts the practice reward parameters to reduce the extrinsic match reward loss computed from matches. We illustrate the method on a simple grid world, and evaluate it in two games in which the practice environment differs from match: Pong with practice against a wall without an opponent, and PacMan with practice in a maze without ghosts. The results show gains from learning in practice in addition to match periods over learning in matches only.
\end{abstract}

\section{Introduction}

There are many applications of reinforcement learning (RL) in which the natural formulation of the reward function gives rise to difficult computational challenges, or in which the reward itself is unavailable for extended periods of time or is difficult to specify. These include settings with very sparse or delayed reward,  multiple tasks or goals, reward uncertainty, and learning in the absence of reward or in advance of unknown future reward.  A range of approaches address these challenges through \emph{reward design}, providing {\em intrinsic} rewards to the agent that augment or replace the objective or {\em extrinsic} reward. The aim is to provide useful and proximal learning signals that drive behavior and learning in a way that improves performance on the main objective of interest \cite{ng1999policy,barto2004intrinsically,singh2010intrinsically}.
These intrinsic rewards are often hand-engineered, and based on either task-specific reward features developed from domain analysis, or task-general reward features, sometimes inspired by intrinsic motivations in animals and humans \cite{Oudeyer,schmidhuber2010formal} and sometimes based on heuristics such as learning diverse skills \cite{gupta2018}. The optimal rewards framework \cite{singh2010intrinsically} provides a general meta-optimization formulation of intrinsic reward design, and has served as the basis for algorithms that discover good intrinsic rewards; we discuss this further in Related Work.

In this work we address the challenges imposed by settings where a learning agent faces extended periods of no evaluation in which an extrinsic reward is unavailable and where the environment may differ from that of objective evaluation when extrinsic reward is available. We refer to such settings as {\em practice-match}, drawing an analogy to regimes of skill acquisition typical for humans in sports and games. For example, in team sports such as basketball it is common to practice skills such as dribbling and shooting in the absence of other players, and in sports such as tennis it is common to practice skills in environments other than a full court. In such settings, during practice, the agent must behave in the absence of the main match reward (e.g., winning games against opponents), but in such a way that performance on the future matches (defined by the extrinsic rewards during match) improves. 
Examples of practice-match settings beyond sports include an office robot using the evening after office-hours to practice for day-time tasks (match), household robotic assistants using free-time to practice, task-specific dialogue agents using down-time to practice with human-trainers or using opportunities for low-stakes on-line conversation practice, and multi-agent teams using down-time to practice coordination strategies.

We focus on the question of how an agent should practice given a practice environment in a setting of alternating periods of practice and match.
We formulate this problem as one of {\em discovering good practice rewards}.
Our primary contribution is a method that learns intrinsic reward functions for practice that improve the match policy during practice. The method uses meta-gradients to adapt the intrinsic practice reward parameters to reduce the extrinsic loss computed from matches. Our results show gains from learning in practice in addition to match periods over the performance achieved from learning in matches only.

\section{Related work}

We place our contributions in the context of three bodies of related work: (a) the design or discovery of intrinsic rewards that modify or replace an available extrinsic reward; (b) the design or discovery of intrinsic rewards to motivate learning and behavior in the absence of extrinsic reward; and (c) meta-gradient approaches to optimizing reinforcement learning agent parameters. 

{\bf Optimal rewards and reward design.}  Reward functions serve as implicit specifications of desired policies, but the precise form of the reward also has consequences for the sample (and computational) complexity of learning. Approaches to \emph{reward design} seek to modify or replace the extrinsic reward to improve the complexity of learning while still finding good policies.  Approaches such as  \emph{potential rewards}  \cite{ng1999policy} define a space of reward transformations guaranteed to preserve the implicit optimal policies. \emph{Intrinsically-motivated RL} aims to improve learning by providing reward bonuses, e.g., to motivate effective exploration, often through hand-designed features that formalize notions such as curiosity or salience \cite{barto2004intrinsically,Oudeyer,schmidhuber2010formal}. In contrast to this prior work, the practice reward discovery method proposed here does not commit to the form of the intrinsic reward and does not use hand-designed reward features. The \emph{optimal rewards} framework of \citeauthor{singh2010intrinsically} \shortcite{singh2010intrinsically} formulates a meta-optimization problem motivated by the insight that the optimal intrinsic reward for an RL agent depends on the bounds on the agent's learning algorithm and environment; algorithms exist for finding optimal intrinsic rewards for planning  \cite{guo2016deep,sorg2010reward} and policy-gradient agents \cite{Zheng2018intrinsic}. Our new work shares the meta-optimization framework of optimal rewards, but addresses the challenge of how to drive learning during periods of practice where extrinsic rewards are not available and the practice environment is different from the evaluation environment.

{\bf Learning in the absence of extrinsic reward.} Recent work addresses the challenge faced by agents that must learn during a period of free exploration that precedes an objective evaluation in which the agent is tasked with a sequence of goals drawn from some distribution; the distribution parameters may be partially known to the agent in advance. 
This prior work includes methods for learning goal-conditioned policies via the automatic generation of a curriculum of goals \cite{Held2018automaticgoals} or via information-theoretic loss functions \cite{eysenbach2018diversity,Goyal2019InfoBotTA}. \citeauthor{gupta2018} \shortcite{gupta2018} generate tasks that lead to learning of diverse skills and use them to learn a policy initialization that adapts quickly to the objective evaluation.  Our work shares with these approaches the challenge of motivating learning in the absence of extrinsic rewards, but differs in that our proposed practice reward method \emph{discovers} intrinsic rewards through losses defined only in terms of an extrinsic reward, and the practice-reward setting concerns a single objective task and possibly different environments.

{\bf Meta-gradient approaches to optimizing RL agent parameters.} 
Recently, researchers have developed several different meta-gradient approaches that optimize meta-parameters of a policy-gradient agent that affect the policy loss only indirectly through their effect on the policy parameters. 
For example, meta-gradient approaches have been used successfully to learn good policy network initializations  that adapts quickly to new tasks \cite{finn2017model,proximal_meta_policy,pmlr-v78-finn17a,gupta2018}, and RL hyper-parameters such as discount factor and bootstrapping parameters \cite{NIPS2018_7507}.  \citeauthor{Zheng2018intrinsic} \shortcite{Zheng2018intrinsic} developed a meta-gradient algorithm for discovering optimal intrinsic rewards for policy gradient agents. Our proposed method modifies and extends \citeauthor{Zheng2018intrinsic} \shortcite{Zheng2018intrinsic} to practice-match settings. Specifically, we derive the gradient of extrinsic reward loss during match with respect to practice reward parameters and use it to improve practice rewards over the course of alternating practices and matches. The success of the method thus contributes to the growing body of recent work demonstrating the utility of meta-gradient algorithms for RL. 

\section{Algorithm for learning practice rewards}

In this section, we first describe briefly policy gradient-based RL and then our algorithm for learning practice rewards.

{\bf Policy gradient- based RL.}~
At each time step $t$, the agent receives a state $s_t$ and takes an action $a_t$ from a discrete set $A$ of possible actions. The actions are taken following a policy $\pi$ (a mapping from states $s_t$ to actions $a_t$), parameterized by $\theta$ and denoted as $\pi_\theta$. The agent then receives the next state $s_{t+1}$ and a scalar reward $r_t$. This process continues until the agent reaches a terminal state (which ends an episode) after which the process restarts and repeats.

Let $G(s_t, a_t)$ be the future discounted sum of rewards obtained by the agent until termination, i.e.,  $G(s_t,a_t) = \sum_{i = t}^{\infty} \gamma^{i-t} r(s_i, a_i)$, where $\gamma$ is the discount factor. 
The value of the policy $\pi_\theta$ denoted by $J(\theta)$ is the expected discounted sum of rewards obtained by the agent when executing actions following the policy $\pi_\theta$, i.e., 
$J(\theta) = E_{\theta} [\sum_{t=0}^{\infty} \gamma^t r(s_t, a_t)]$.
The policy gradient theorem of ~\citeauthor{sutton2000policy} \shortcite{sutton2000policy} shows that for all time steps $t$ within an episode, the gradient of the value $J(\theta)$ with respect to the policy parameters $\theta$ can be obtained as follows:
\begin{align}
\nabla_{\theta}J(\theta) =  E_{\theta}[G(s_t,a_t) \nabla_{\theta}\log \pi_\theta(a_t|s_t)]
\label{eq:pg}
\end{align}

{\bf Notation.}~ We use the following notation throughout:

\begin{center}
    \begin{itemize}
      \item $\theta$ : policy parameters 
    \item $r^{ex} = r^{ex}(s,a)$ : extrinsic reward (available during matches) 
    \item $r_\eta^{in} = r_\eta^{in} (s,a)$ : intrinsic reward parameterized by $\eta$ 
     \item $G^{ex} (s_t, a_t) = \sum_{i=t}^{\infty} \gamma^{i-t}r^{ex}(s_i, a_i)$ : extrinsic reward return 
    \item $G^{in} (s_t, a_t) = \sum_{i=t}^{\infty} \gamma^{i-t}r_\eta^{in} (s_i, a_i)$ : intrinsic reward return 
     \item $J^{ex} = E_{\theta}[\sum_{t=0}^{\infty} \gamma^t r^{ex}(s_t, a_t)]$ : extrinsic value of policy 
     \item $J^{in} = E_{\theta}[\sum_{t=0}^{\infty} \gamma^t r^{in}_\eta(s_t, a_t)]$ : intrinsic value of policy 
    \end{itemize}
\end{center}

\begin{figure}
    \centering
    \includegraphics[width=0.72\columnwidth]{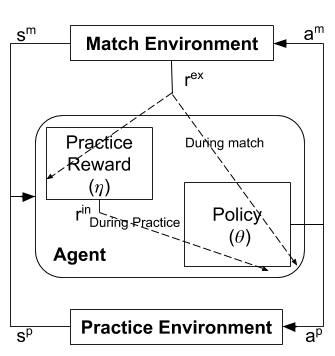}
    \caption{The agent has two modules, the policy module parameterized by $\theta$ and the practice reward module parameterized by $\eta$. As the dashed lines show, $\theta$ is updated using extrinsic match reward $r^{ex}$ during match and using the intrinsic practice reward $r^{in}$ during practice; $\eta$ is updated using the extrinsic match reward $r^{ex}$ from the match environment.}
    \label{fig:arch}
\end{figure}

{\bf Algorithm overview.}~
The algorithm is specified in Algorithm \ref{alg:practice_ai} and the agent architecture is depicted in Figure~\ref{fig:arch}. At each time step $t$ the agent receives an observation from the environment and \textit{concatenates the observation with a practice/match flag} indicating whether the agent is in practice or match. We denote this concatenated input as $s_t^p$ for the practice environment and $s_t^m$ for the match environment.

During {\em match}, the policy parameters are updated to improve performance in the match task as defined by the extrinsic reward; this happens by adjusting the policy parameters $\theta$  in the direction of the gradient of $J^{ex}$,  which is the expected discounted sum of match time extrinsic rewards.

During {\em practice}, the policy parameters are updated to improve performance in the practice task as defined by the current intrinsic practice reward; this happens by adjusting $\theta$ in the direction of the gradient of $J^{in}$,  which is the expected discounted sum of practice time intrinsic rewards. 

After each practice update, the intrinsic practice reward parameters are updated in the key meta-gradient step. The aim is to adjust the intrinsic practice reward so that the policy parameter updates that result from practice improve the extrinsic reward performance on match. This is done by using match experience to evaluate the policy parameters that result from the practice update, and updating the intrinsic reward parameters $\eta$ in the direction of the gradient of $J^{ex}$ computed on the match experience. We explore  two variants: updating based on the previous match experience, and updating based on the next match experience. We describe each step in detail below. 

Our algorithm is a modification and extension of \citeauthor{Zheng2018intrinsic} \shortcite{Zheng2018intrinsic}'s algorithm (which discovers optimal intrinsic rewards for policy gradient agents in the regular RL setting) for practice-match settings and we follow their derivations closely.

{\bf Updating policy parameters during match.}~
Let $\mathcal{D}_m = \{ s_0^m, a_0^m, s_1^m, a_1^m, \cdots \}$ be the trajectory taken by the agent in the match using the policy $\pi_\theta$. The policy parameters $\theta$ are updated in the direction of the gradient of $J^{ex}$: 
\begin{align}
    \theta' &=  \theta + \alpha^m \nabla_{\theta} J^{ex}(\theta;\mathcal{D}_m)\\
    & \approx \theta + \alpha^m G^{ex}(s_t^m, a_t^m) \nabla_{\theta} \log \pi_{\theta} (a_t^m | s_t^m)
    \label{eqn:match_policy_update}
\end{align}
using the empirical return $G^{ex}$ in the approximation of the gradient.

{\bf Updating policy parameters during practice.}~
Let $\mathcal{D}_p = \{ s_0^p, a_0^p, s_1^p, a_1^p, \cdots \}$ be the trajectory taken by the agent in the practice environment using the policy $\pi_{\theta'}$. The policy parameters $\theta'$ are updated in the direction of the gradient of $J^{in}$:
\begin{align}
    \theta'' &=  \theta' + \alpha^p \nabla_{\theta'} J^{in}(\theta';\mathcal{D}_p)\\
    & \approx \theta' + \alpha^p G^{in}(s_t^p, a_t^p) \nabla_{\theta'} \log \pi_{\theta'} (a_t^p | s_t^p)
    \label{eqn:practice_policy_update}
\end{align}
using the empirical return $G^{in}$ in the approximation of the gradient.

{\bf Updating intrinsic practice reward parameters.}~
The intrinsic practice reward parameters $\eta$ are updated in the direction of the gradient of $J^{ex}$ of the match. The gradient of $J^{ex}$ of the match with respect to $\eta$ is computed  using the chain rule as follows:
\begin{align}
    \eta' &=  \eta + \beta \nabla_{\eta} J^{ex}(\theta'')\\
     &= \eta + \beta \nabla_{\eta} \theta'' \nabla_{\theta''} J^{ex}(\theta'')
     \label{eqn:intrinsic_update}
\end{align}
The second term $\nabla_{\theta''} J^{ex}(\theta'')$ evaluates the policy parameters $\theta''$ (that resulted from the practice update using the intrinsic rewards) using  match samples. We specify here two forms of the intrinsic practice reward update: when the match samples are from the {\em next} match, and when the match samples are from the {\em previous} match.
If we use the next match to perform the update, the agent will act using the policy $\pi_{\theta''}$ in the next match and can use the new match samples from the trajectory $\mathcal{D}_{m'}$ to approximate $\nabla_{\theta''} J^{ex}(\theta'')$ as follows: 
 \begin{align}
     \nabla_{\theta''} J^{ex}(\theta'') &= \nabla_{\theta''} J^{ex}(\theta'';\mathcal{D}_{m'}) \notag \\ 
     &\approx G^{ex} (s_t^{m'}, a_t^{m'}) \nabla_{\theta''} \log \pi_{\theta''} (a_t^{m'} | s_t^{m'}) \label{eqn:intrinsic_update_1a}
\end{align}
If we use the previous match samples, the agent can perform an off-policy update using an importance sampling correction:
\begin{align}
    \nabla_{\theta''} J^{ex}(\theta'') &= \nabla_{\theta''} J^{ex}(\theta'';\mathcal{D}_m) \notag \\ 
    &\approx G^{ex} (s_t^m, a_t^m) \frac{\nabla_{\theta''} \log \pi_{\theta''} (a_t^m | s_t^m)}{ \log \pi_{\theta} (a_t^m | s_t^m)} \label{eqn:intrinsic_update_1b}
\end{align}
The first term $\nabla_{\eta} \theta''$ in Eq. \ref{eqn:intrinsic_update} evaluates the effect of change in the intrinsic parameters $\eta$ on the policy parameters that result after the practice time policy update, $\theta''$. This term can be computed as follows:
\begin{align}
    \nabla_{\eta} \theta'' &= \nabla_{\eta} \left(\theta' + \alpha^p G^{in}(s_t^p, a_t^p) \nabla_{\theta'} \log \pi_{\theta'} (a_t^p | s_t^p)\right)\\
    &= \nabla_{\eta} \left(\alpha^p G^{in}(s_t^p, a_t^p) \nabla_{\theta'} \log \pi_{\theta'} (a_t^p | s_t^p)\right)\\
    &= \nabla_{\eta} \left(\alpha^p (\sum_{i=t}^{\infty} \gamma^{i-t}r_\eta^{in} (s_i^p, a_i^p)) \nabla_{\theta'} \log \pi_{\theta'} (a_t^p | s_t^p)\right)\\
    &= \alpha^p \left(\sum_{i=t}^{\infty} \gamma^{i-t} \nabla_{\eta} r_\eta^{in} (s_i^p, a_i^p)\right) \nabla_{\theta'} \log \pi_{\theta'} (a_t^p | s_t^p)
    \label{eqn:intrinsic_update_2}
\end{align}

\begin{algorithm}[tb]
\caption{Learning Practice Rewards}
\begin{algorithmic}[1]
	\STATE{\bfseries Input:} step-size parameters $\alpha^m$,  $\alpha^p$ and $\beta$
    \STATE{\bfseries Init:} initialize $\theta$ and $\eta$ with random values
    \REPEAT
        \STATE {\bf Updating policy parameters during match}
    	\STATE  Sample a trajectory $\mathcal{D}_m=\{s_{0}^m, a_{0}^m, s_{1}^m, a_{1}^m, \cdots\}$ by interacting with the match environment using $\pi_\theta$
        \STATE  Approximate $\nabla_{\theta} J^{ex}(\theta;\mathcal{D}_m)$ by Equation \ref{eqn:match_policy_update}
        \STATE  Update $\theta' \leftarrow \theta + \alpha^m \nabla_{\theta}J^{ex}(\theta;\mathcal{D}_m)$
        
        \STATE {\bf Updating policy parameters during practice}
    	\STATE  Sample a trajectory $\mathcal{D}_p=\{s_{0}^p, a_{0}^p, s_{1}^p, a_{1}^p, \cdots\}$ by interacting with the practice environment using $\pi_{\theta'}$
        \STATE  Approximate $\nabla_{\theta'} J^{in}(\theta';\mathcal{D}_p)$ by Equation \ref{eqn:practice_policy_update}
        \STATE  Update $\theta'' \leftarrow \theta' + \alpha^p \nabla_{\theta'}J^{in}(\theta';\mathcal{D}_p)$
        \STATE {\bf Updating intrinsic reward parameters after practice update}
        \STATE  Approximate $\nabla_{\theta''} J^{ex}(\theta''$) by Equation \ref{eqn:intrinsic_update_1a} or \ref{eqn:intrinsic_update_1b}
        \STATE  Approximate $\nabla_{\eta} \theta''$ by Equation \ref{eqn:intrinsic_update_2}
        \STATE  Compute $\nabla_{\eta} J^{ex} = \nabla_{\theta''} J^{ex}(\theta'') \nabla_{\eta} \theta''$
        \STATE  Update $\eta' \leftarrow \eta + \beta \nabla_{\eta} J^{ex}$
    \UNTIL{done}
\end{algorithmic}
\label{alg:practice_ai}
\end{algorithm}

For simplicity we have described our proposed algorithm using a basic policy gradient formulation. Our proposed algorithm is fully compatible with  advanced policy gradient methods such as Advantage Actor-Critic that reduce the variance of the gradient and improve data efficiency.

\section{Illustration on grid-world: \\ Visualizing practice rewards}

\begin{figure*}
\begin{subfigure}{0.5\columnwidth}
    \centering
    \includegraphics*[width=0.7\columnwidth]{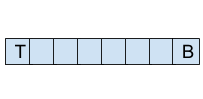}
    \caption{Corridor world}
    \label{fig:corridor}
\end{subfigure}
\begin{subfigure}{0.5\columnwidth}
    \centering
    \includegraphics[width=\columnwidth]{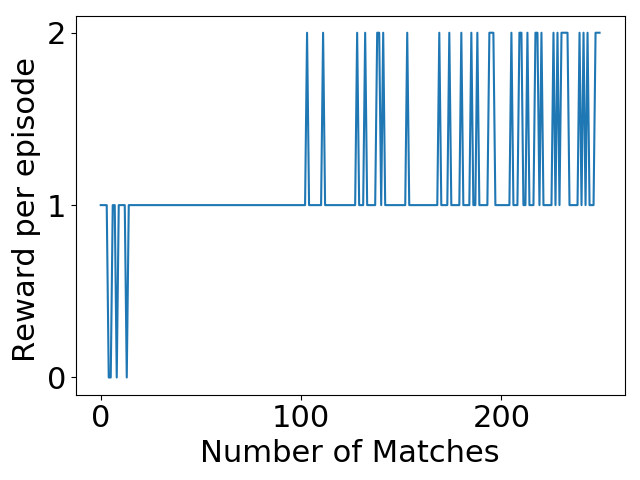}
    \caption{}
    \label{fig:corridor_1}
\end{subfigure}
\begin{subfigure}{0.5\columnwidth}
    \centering
    \includegraphics[width=\columnwidth]{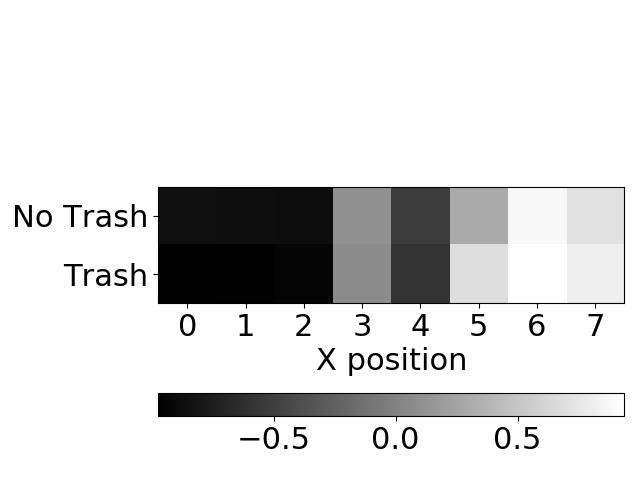}
    \caption{Match 1}
    \label{fig:corridor_viz_1}
\end{subfigure}
\begin{subfigure}{0.5\columnwidth}
    \centering
    \includegraphics[width=\columnwidth]{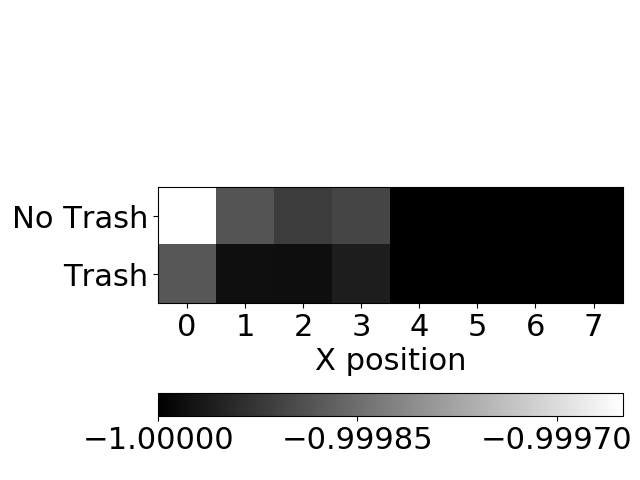}
    \caption{Match 200}
    \label{fig:corridor_viz_2}
\end{subfigure}
\caption{{\bf (a)} shows the corridor world with trash (T) in the left most corner and the bin (B) in the rightmost corner. {\bf (b)} shows the learning curve of the agent in the corridor world. The x-axis is number of matches during learning. The y-axis is the reward per episode during matches. {\bf (c)} and {\bf (d)} show the visualization of the learned intrinsic rewards for practice over the state space at two different points in the agent's learning. The top and bottom rows correspond to the agent carrying trash and not carrying trash respectively. The visualization maps the intrinsic reward values to the lightness of the color with dark (black) corresponding to the lowest value and fully illuminated (white) corresponding to the highest.
The corresponding color bars show what exact value a color represents.}
\label{fig:corridor_full}
\end{figure*}

We now illustrate the algorithm in a simple grid world that allows us to visualize discovered practice rewards at different points in the agent's learning. The environment is a corridor world of length 8 shown in Figure \ref{fig:corridor}.
The corridor world has {\em trash} (T) in the leftmost corner $(X = 0)$ and a {\em bin} (B) in the rightmost corner $(X = 7)$. The state input for the agent is its $X$ position, a flag denoting if it has trash or not and flag denoting if it is in practice or match. The agent has two actions, move left and move right. The agent starts every episode at $X = 0$ with trash. If the agent moves to the bin, $X = 7$, with trash it gets a reward of $+1$ for delivering the trash and it automatically loses the trash at the following time step. If it moves back to $X = 0$ without trash, it gets the trash automatically at the following time step.
The agent undergoes 3 practice episodes before every match episode. Here, the match and the practice environment are the same. Each episode in both practice and match is of length between 45 and 50, sampled uniformly. The agent uses REINFORCE \cite{williams1992simple} with our proposed algorithm for its learning. Next match samples are used for updating the intrinsic practice reward parameters using Equation \ref{eqn:intrinsic_update_1a}. More details on the architecture and training are provided in the Appendix.

Intuitively there are two important stages in the learning for this task. First, the agent must learn to take the trash from $X=0$ to $X=7$.  Second, the agent must learn to come back to $X=0$ to collect the trash again, so that the first step can be repeated. Figure \ref{fig:corridor_1} shows the return obtained by the agent across the matches. We observe that the agent quickly learns to get a episode reward of $+1$ and later, after about 100 matches, starts getting a episode reward of $+2$. 

{\bf Visualization of learned intrinsic practice rewards.}~ 
Our aim here is to visualize how good practice rewards vary as a function of the learning state of the agent. We do this by pausing the update of the policy at two different points during learning (Match 1 and Match 200),  and allowing the intrinsic reward parameters to be updated (via additional samples of match and practice experience) until they converge. In other words, we are seeking to visualize an approximation of the optimal practice reward as a function of learning.  (To be clear, the results in Figure~\ref{fig:corridor_1} are from Algorithm \ref{alg:practice_ai} without pausing to allow intrinsic reward convergence.)

Figure \ref{fig:corridor_viz_1} shows the (approximate) optimal practice reward over the state space at the start of agent's learning (Match 1). The top and bottom rows correspond to the agent carrying trash and not carrying trash respectively. The reward tends to be high (darker) towards the right and low (lighter) in the left of the corridor (irrespective of the presence or absence of trash), which indicates that it is asking the agent to practice going from left to right, which would allow it to get an extrinsic reward of $+1$ during match, as the agent always begins an episode at the leftmost corner with trash. 
Figure \ref{fig:corridor_viz_2} shows the (approximate) optimal practice reward for an agent that has learned over 200 matches. At this point the agent consistently gets a reward of at least $+1$ (see Figure~\ref{fig:corridor_1}), which means starting from $X=0$ with trash at the beginning of the episode, the agent has learned to take the trash to $X=7$ (bin) once.
Now it needs to learn to go back to $X=0$ from $X=7$ (bin), so that it can collect the trash, and take it to the bin again to get an additional reward of $+1$. Figure \ref{fig:corridor_viz_2} indicates that the (approximate) optimal practice reward encourages such behavior in practice. In order to reach the highest rewarding state of $X = 0$ and No Trash, the agent which starts at $X=0$ with trash has to go to the bin, $X=7$ (where it loses the trash) and come back to $X=0$. In the following time step, it will automatically get trash. Now the agent has to repeat the above to reach the highest rewarding state ($X=0$, No Trash) again, which leads to the desired behavior of repeatedly collecting and emptying the trash.

These visualizations show that our meta-gradient learning method finds practice rewards that have an intuitive and expected interpretation in this simple domain, and furthermore they highlight an important (and understudied) aspect of learning intrinsic rewards in general: that good intrinsic rewards are non-stationary because they depend on the state of the learner.
We now move to evaluations in more challenging domains in which practice and match environments differ.

\begin{figure*}
\begin{subfigure}{0.4\columnwidth}
\begin{subfigure}{\columnwidth}
    \centering
    \includegraphics[width=\columnwidth]{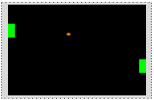}
    \caption{Match env}
    \label{fig:pong_match}
\end{subfigure}
\begin{subfigure}{\columnwidth}
    \centering
    \includegraphics[width=\columnwidth]{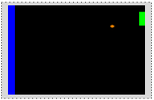}
    \caption{Practice env}
    \label{fig:pong_practice}
\end{subfigure}
\end{subfigure}
\begin{subfigure}{0.8\columnwidth}
    \centering
    \includegraphics[width=\columnwidth]{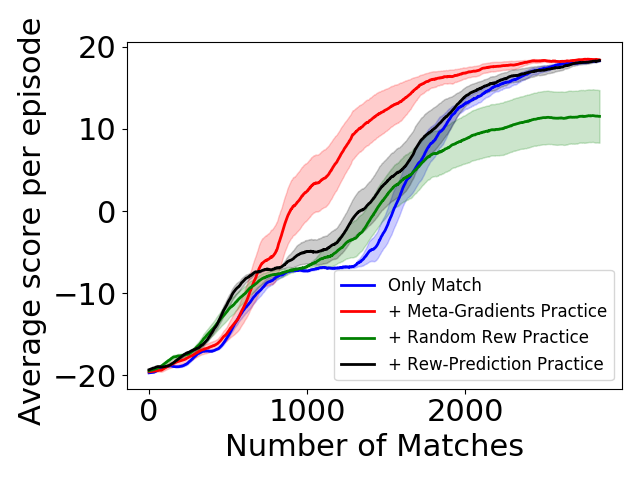}
    \caption{}
    \label{fig:pong_xmatch}
\end{subfigure}
\begin{subfigure}{0.8\columnwidth}
    \centering
    \includegraphics[width=\columnwidth]{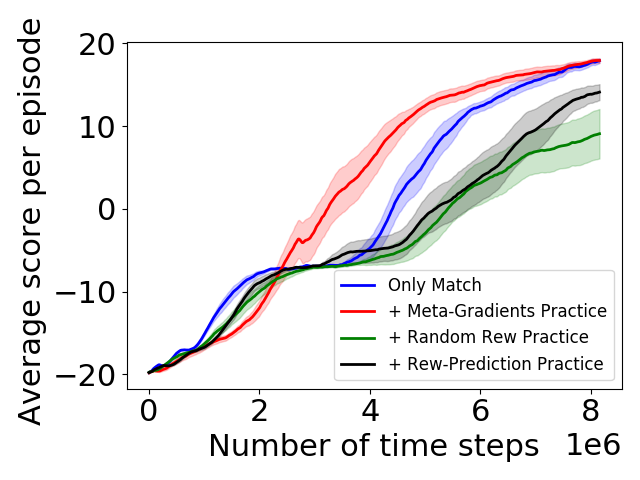}
    \caption{}
    \label{fig:pong_xtime}
\end{subfigure}
\caption{{\em Results from Pong.} The blue, red, green and black curves show, respectively, performance for the baseline A2C agent learning only in matches, the practicing A2C agent using meta-gradient updates to improve the practice reward, the practicing A2C agent using fixed random rewards and, the practicing A2C agent using rewards from the extrinsic reward prediction network.
The curves are the average of $10$ runs with different random seeds, the shaded area shows the standard error.
The y-axis is the mean  reward over the last $100$ training episodes. For {\bf (c)} the x-axis is the number of matches during learning and for  {\bf (d)} the x-axis is the number of time steps during learning in both practice (when performed) and match combined.}
\label{fig:pong}
\end{figure*} 

\section{Evaluation on practice-match versions of two Atari games}
In the following two experiments we create practice-match settings of two Atari games in which the practice environment differs from the match environment in an interesting way.  We perform comparisons to baseline conditions to answer the following questions:
\begin{enumerate}
    \item Does learning in practice environments in addition to matches improve performance compared to learning in matches only?
   \item Is the meta-gradient update for improving the practice reward contributing to performance improvement above that obtained from training with a fixed random practice reward?
   \item How does the proposed meta-gradient based method for learning practice rewards compare with a method that provides practice rewards that are similar to the match time extrinsic rewards?
    \item How does the performance obtained from practice and match compare with the performance obtained if the time allotted to practice was instead replaced with additional matches?
\end{enumerate}
To answer the first and fourth questions we measure and report on the comparisons therein below.
To answer the second question we initialize the practice reward parameters with random weights using the same initialization method as in the meta-gradient agents, but we keep the practice reward parameters fixed during learning. In this way we directly test the effect of the meta-gradient update.
To answer the third question, we design a method where the intrinsic rewards used during practice come from a network that is trained to predict extrinsic rewards during matches. This is a sensible approach to learning potentially useful practice rewards and may be very effective in certain practice-match settings.
 
The two domains used for our evaluation are Pong and PacMan. In Pong, the practice environment has a wall on the side opposite to the agent instead of an opponent.  In PacMan, the practice environment has the same maze as match but without any ghosts (ghosts are other agents that must be avoided). After every match, the agent is allowed a fixed time for practice in its practice environment.

{\bf Implementation details.}~ The learning agent uses the open-source implementation of the A2C algorithm \cite{mnih2016asynchronous} from OpenAI \cite{baselines} for the two games. A2C performs multiple updates to the policy parameters within a single episode (both in practice and match). Instead of waiting for the next match, we store the previous match samples in a buffer and use them to evaluate the practice policy updates as they happen within a practice episode and update the intrinsic reward parameters.
The extrinsic reward provided to the agent during match is the change in game score as is standard in work on Atari games. The image pixel values and the practice/match flag are provided as state input to the A2C agent (policy and the practice reward modules). The practice reward module outputs a single scalar value (through a tanh non-linearity). More details on architecture and training are provided in the Appendix.
There is a visual mismatch between the practice and match environments (described below) which the agent must learn to account for while transferring learning from practice to match. Note that the agent has the information of whether it is in practice or match as a part of its state input which enables the agent to learn different policies for practice and for match.

For both Pong and PacMan, we show learning curves for four A2C agents: an A2C agent that learns only in matches, an A2C agent that learns in both practice and match using our new algorithm (+ Meta-Gradients Practice), an A2C agent that learns in practice and match but using a fixed random practice reward network during practice (+ Random Rew Practice) and an A2C agent that learns in practice and match but using the practice rewards during practice from a network that is trained to predict extrinsic rewards during matches (+ Rew-Prediction Practice).

{\bf Pong experiments.}~ Pong is a two player game that simulates table tennis. Each player controls a paddle which can move vertically to hit a ball back and forth. The RL agent competes against a CPU player on the opposite side. The goal is to reach twenty points before the opponent does; a point is earned when the opponent fails to return the ball. The dynamics are interesting in that the return angle and speed of the ball depends on where the ball hits the paddle.

In the {\em practice environment} there is no opponent but instead a wall on the opponent's side can bounce the ball back. In contrast to an opponent's paddle, the angle of rebound is always the same as the angle of incidence irrespective of where the ball hits the wall, and the acceleration remains constant as well. Figures \ref{fig:pong_match} and \ref{fig:pong_practice} show the match and practice environments.

To perform well in Pong, the agent needs to learn to track the ball and return it to the opponent so that the opponent misses it. This requires the agent to use the opponent's location to determine where on the paddle the ball should be hit to control the return direction and speed of the ball.
The practice environment potentially allows the agent to practice tracking and returning the ball successfully without missing it, but it does not help prepare the agent for the varying speeds and direction of the ball when returned from an opponent's paddle. The practice environment also does not help practicing for directing the return of ball depending on the opponent's position.  
The agent practices in this modified practice environment for 3000 time steps after every match.

{\bf PacMan experiments.}~The player moves a PacMan through a maze containing stationary pellets and moving ghosts. The player earns points by eating pellets;  the goal is to eat as many pellets as possible while avoiding the ghosts. There are two power pellets that provide a temporary ability to eat ghosts and earn bonus points. The match ends if the PacMan eats all the pellets, the PacMan is eaten by the ghost, or the number of time steps reaches the limit of 200.

\begin{figure*}
\begin{subfigure}{0.4\columnwidth}
\begin{subfigure}{\columnwidth}
    \centering
    \includegraphics[width=\columnwidth]{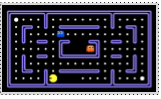}
    \caption{Match env}
    \label{fig:PacMan_match}
\end{subfigure}\hfill
\begin{subfigure}{\columnwidth}
    \centering
    \includegraphics[width=\columnwidth]{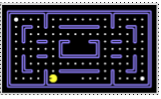}
    \caption{Practice env}
    \label{fig:PacMan_practice}
\end{subfigure}
\end{subfigure}
\begin{subfigure}{0.8\columnwidth}
    \centering
    \includegraphics[width=\columnwidth]{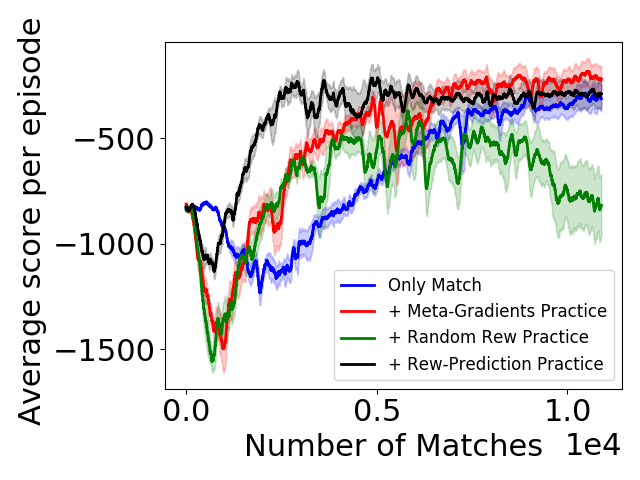}
    \caption{}
    \label{fig:PacMan_performance_match}
\end{subfigure}
\begin{subfigure}{0.8\columnwidth}
    \centering
    \includegraphics[width=\columnwidth]{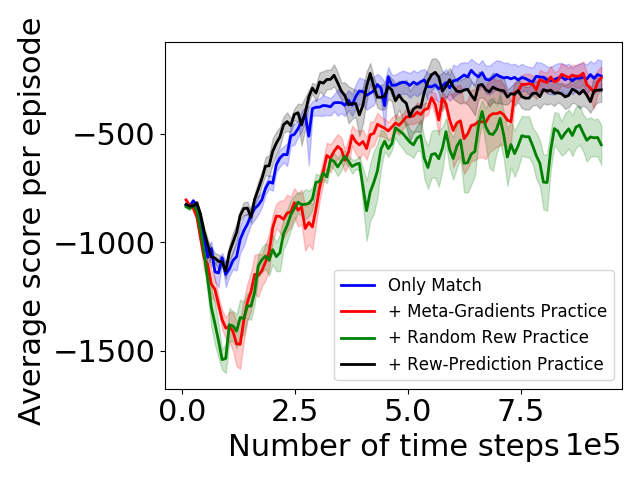}
    \caption{}
    \label{fig:PacMan_performance_time}
\end{subfigure}

\caption{{\em Results from PacMan.} The blue, red, green and black curves show, respectively, performance for the baseline A2C agent learning only in matches, the practicing A2C agent using meta-gradient updates to improve the practice reward, the practicing A2C agent using fixed random rewards and, the practicing A2C agent using rewards from the extrinsic reward prediction network. The curves average $10$ runs with different random seeds, the shaded area shows the standard error. The y-axis is the mean reward over the last $100$ training episodes. For {\bf (c)} the x-axis is the number of matches during learning and for  {\bf (d)} the x-axis is the number of time steps during learning in both practice (when performed) and match combined.}

\label{fig:PacMan}
\end{figure*}

The {\em practice environment} has the same maze with pellets, but does not have any ghosts (Figs. \ref{fig:PacMan_match} and \ref{fig:PacMan_practice}). Each practice episode lasts 100 time steps, and there are 
3 practice episodes after every match.
To perform well in a PacMan match, the agent must learn to identify where pellets are in the maze and navigate to them efficiently, while avoiding ghosts and taking alternate routes when needed. The practice environment allows the agent to learn to navigate the maze to eat pellets but does not allow it to learn to avoid ghosts and take alternate routes depending on the ghost's position during the process of trying to eat the pellets.

{\bf Pong and PacMan results.}~ 
Figures \ref{fig:pong_xmatch} and \ref{fig:PacMan_performance_match} show the average score that the four A2C agents obtained per episode across matches in Pong and PacMan respectively. We see that learning in practice periods in addition to match periods using our proposed method (red curve) helps the agent reach good performance faster than just learning in the matches (blue curve), answering our first question above. This question of whether learning in practice in addition to match is helpful, is one that may be of significant applied interest. For example, this question is important in all of our motivating examples: basketball, tennis, or any sports, office robot, household robot, task-specific dialog agent and multi-agent teams. In all these scenarios practice can be done in addition to match without affecting the matches themselves. In other words, removing the practice (which is available in between the matches) will not speed up the availability of matches.

Figures \ref{fig:pong_xmatch} and \ref{fig:PacMan_performance_match} also show clearly that the benefit from practice is due to the meta-gradient update. The agent practicing with a fixed random intrinsic practice reward (green curve) performs very poorly compared to the method that improves the intrinsic practice rewards using meta-gradient updates (red curve). This answers our second question.

The black curve (Rew-Prediction practice) shows the performance of the method where the intrinsic rewards used during practice come from a network that is trained to predict extrinsic match rewards during matches. This is a sensible approach to learning potentially useful practice rewards and may be very effective in certain practice-match settings such as in PacMan, where we expect it would provide practice rewards for eating pellets, a very good practice reward for practice without ghost. 

In Pong this baseline performs worse than our proposed method for learning intrinsic rewards. In PacMan, in the initial stages of learning, this baseline provides much faster learning compared to our proposed method.
However it ends up settling to a solution which is slightly worse than our proposed method. This is an interesting outcome because it suggests that, though it takes some time to learn the intrinsic practice rewards, our method can learn better practice rewards. We conjecture that this is because our method can adapt practice reward across the agent's lifetime and exploit the capacity to take into consideration how policy parameter changes during practice affect the match time policies which the baseline method cannot do. Further study is required to understand when our proposed method based on meta-gradients can provide faster learning compared to Rew-Prediction practice. This might be closely tied to the question of how the relationship between practice and match environment impact the performance of the two methods. This answers our third question.

Figures \ref{fig:pong_xtime} and \ref{fig:PacMan_performance_time} show learning curves as a function of {\em time steps} in {\em both} practice and match combined.
This compares the performance of an agent that learns in practice and match with that of an agent whose practice time is replaced with additional matches (blue curve). In other words, it answers question 4.
Surprisingly in Pong the agent could learn to perform better in matches faster if it uses some time on practice in the modified environment---while learning practice rewards using our proposed method (red curve)---instead of using that time playing additional matches.
Whether it is  possible to achieve faster and better learning in matches through practices instead of additional matches depends on how the practice environment is related to that of the match. In PacMan where the match policy is highly dependent on ghost position, practice without ghosts may not substitute for additional matches even if the agent performs the best practice possible. This is reflected in the results as well. In Pong we hypothesize that  practice with a wall is an easier environment to learn returning the ball compared to a match with an opponent and hence leads to faster learning compared to having additional matches. 

However in both Pong and PacMan, as we have seen, when we have practice in addition to matches, it leads to faster learning for a given number of matches compared to learning in matches only. 
{As noted earlier, this evaluation of performance with respect to the number of matches is one of practical interest.

\section{Conclusion}

In this work we address the challenges encountered when a learning agent must learn in an environment in which the extrinsic reward of a primary task is not available, and where the environment itself may differ from the primary task environment; the {\em practice-match} setting.
To address these challenges we formulated a {\em practice reward discovery} problem and proposed a principled meta-gradient method to solve the problem. We provided evidence from a simple grid world that shows that good practice rewards discovered by the method depend on the state of the learner. 

In our primary evaluations on Pong and Pacman the practice environments differed from the standard match environments. The performance obtained from practicing in addition to match exceeded that in match alone, even though the agent had to learn {\em what} it should practice---that is, learn the practice reward---in addition to learning to improve the policy on the match task through the practice itself. The comparison to a poorly-performing fixed random practice reward provided evidence that performance gains are due to the meta-gradient update of the practice reward.

Conclusions concerning the generality of the method are limited by the properties of our present evaluations. We do not yet know how effective the method will be when combined with a broader range of agent architectures, although in principle it should be possible to use it with any kind of policy gradient method. The Atari experiments provide some evidence for this in their use of the A2C actor-critic architecture. We also do not yet know how the effectiveness of the method depends on the extent of the difference between match and practice environments. Because the possible benefits of practice are limited by the environment used for practice, an important direction for future work is to understand which environments are well suited for practice and how to construct them, possibly automatically.

More broadly, our results provide additional evidence for the perhaps surprising effectiveness of meta-gradient approaches in reinforcement learning, and more specifically for the effectiveness of methods for adapting rewards.  But like any meta-gradient method that depends on a signal from a primary task gradient, very delayed/sparse and difficult-to-obtain rewards remain significant challenges.  These challenges suggest important directions for future research. 

\section{Acknowledgements}
This work was supported by grants from Toyota Research Institute and from DARPA's L2M program. Any opinions, findings, conclusions, or recommendations expressed here are those of the authors and do not necessarily reflect the views of the sponsors.

\bibliographystyle{aaai}
\bibliography{aaai20}

\appendix
\begin{appendices}
\section{Appendix}
In this section, we describe the details of our learning agents, their environments, and their training.

\subsection{Grid-world experiments}
\label{app:corridor}
{\bf Algorithm.}~The learning of the agent is achieved through REINFORCE~\cite{williams1992simple}. The variance of policy gradients in this REINFORCE agent is reduced using a value function baseline. 

{\bf Agent architecture.}~The agent consists of three neural networks: policy network, value network and the intrinsic practice reward network. The policy and value networks are parameterized by $\theta$; the intrinsic practice reward network that forms the practice reward module is parameterized by $\eta$. Hidden activations are produced using $relu$.

Each of these three networks have 2 hidden layers of 64 units. The policy network has a output layer with dimensions equal to the number of actions in the environment (2 - left and right). The value network maps to a scalar value and the intrinsic network maps to a scalar value with a $tanh$ activation at the end.

{\bf Corridor environment.}~The Corridor world is 1-dimensional grid-world of length $8$, with trash located the leftmost state $X=0$ and bin located at the rightmost state $X=7$. The agent has two actions available: Move left and Move right. If the agent reaches $X=0$, it automatically picks up the trash and if the agent reaches $X=7$ with trash, it automatically deposits the trash in the bin which results in a reward of $+1$ (the extrinsic reward is available only during the matches). The agent receives as state, the concatenation of its $X$ position, a flag indicating if it has the trash and a flag indicating if the agent is in practice or match.  

{\bf Training details.}~Adam optimizer is used for updating the parameters of the learning agent. The initial learning rate  is set to $0.01$. During a practice phase, the agent's parameters are updated on-policy using intrinsic rewards produced by $\eta$. After practice phase, the updated policy parameters $\theta$ are used to act in a match. The trajectory from this match is used to update intrinsic practice reward parameters $\eta$.

\subsection{Atari experiments}
\label{app:atari}
{\bf Algorithm.}~The learning agent is an advantage actor-critic (A2C) architecture~\cite{mnih2016asynchronous} for both Pong and PacMan Atari games. We used the OpenAI baselines~\cite{baselines} for our implementation.

{\bf Agent architecture.}~The learning agent consists of three neural networks and their semantics are same as the one described in our Grid-world experiments.
The policy, value and intrinsic reward networks are convolutional neural networks. The architecture for the networks consists of 3 convolutional layers, followed by a fully-connected layer with $512$ units. The convolution layers consists of 32, 64, 64 filters respectively; the filter-sizes for these layers are $8 \times 8, 4 \times 4, 3 \times 3$ and with stride lengths of $ 4, 2, 1 $ respectively. The activations from the fully-connected layer is then used for producing the network's output. $relu$ activations are used.

The practice/match flag is represented as a one-hot vector, and this is used to produce an embedding of size $512$. This vector is then added to the activations produced by the penultimate fully-connected layer of the policy, value and intrinsic reward networks.

The policy network and the value network share the convolutional layers and the fully-connected layer. The output dimension of the policy network is $|A|$ and value network outputs a scalar value. The intrinsic reward network also outputs a scalar value with $tanh$ activation.

{\bf Pong and PacMan environments.}~The Pong match and practice environments are based on the open-sourced implementation from~\citeauthor{pong}~\shortcite{pong}. The PacMan match and practice environments are designed over the open-sourced implementation available from~\citeauthor{pacman_berkeley}~\citeyear{pacman_berkeley}. 

{\bf Training details.}~We follow the standard pre-processing steps that was introduced in~\citeauthor{mnih2015human}~\shortcite{mnih2015human}. The shape of observations in our Pong environment is $ 84 \times 84 $ and is $ 42 \times 75$ in our PacMan environment. They are grayscale images. The extrinsic rewards from the game are clipped to $[-1,1]$. 

For the policy module, both the baseline agents and our proposed method agent use the default values for all hyper-parameters provided by OpenAI~\cite{baselines} implementation of A2C. For the intrinsic reward module, we use RMSProp for optimization, with a decay factor of $0.99$ and $\epsilon$ $0.00001$. The step size $\beta$ is initialized to $0.0007$ and annealed linearly to zero over the agents learning.

The intrinsic reward parameters are trained in an off-policy manner. Specifically, we use a replay buffer to store samples from matches. After each practice phase, we evaluate the meta-objective (A2C loss function) using the updated policy parameters on a batch of match samples from the replay buffer and compute gradients for the intrinsic reward parameters with this meta-objective.

We performed a hyper-parameter searches for the replay buffer size and batch size used for computing our meta-objective. For our experiments on Pong, we used a buffer size of $12000$ and batch size of $1000$; and for PacMan, we used a buffer size of $4000$ and batch size of $1000$.
\end{appendices}

\end{document}